\documentclass[runningheads,a4paper,orivec]{llncs}

\usepackage[T1]{fontenc}
\usepackage{graphicx}
\usepackage{booktabs}
\usepackage{amsmath,amssymb}
\usepackage{multirow}
\usepackage{xcolor}
\usepackage{tikz}
\usetikzlibrary{positioning,arrows.meta,fit,backgrounds}
\usepackage[pdfusetitle=false]{hyperref}
\hypersetup{
  pdftitle={GAZE: Grounded Agentic Zero-shot Evaluation with Viewer-Level Tools and Literature Retrieval on Rare Brain MRI},
  pdfauthor={Duaa Alim and Mogtaba Alim and Liam Chalcroft},
  hidelinks
}

\newcommand{\framework}{GAZE}

\begin{document}
\raggedbottom

\title{GAZE: Grounded Agentic Zero-shot Evaluation with Viewer-Level Tools and Literature Retrieval on Rare Brain MRI}
\titlerunning{GAZE: Agentic Medical VLM Evaluation on Rare Brain MRI}

\author{Duaa Alim\inst{1}\orcidID{0009-0001-9041-7074} \and
Mogtaba Alim\inst{2} \and
Liam Chalcroft\inst{3}\orcidID{0000-0003-3363-6454}}
\authorrunning{D. Alim et al.}
\institute{Imperial College London, UK \and
University of Toronto, Canada \and
University College London, UK\\
\email{l.chalcroft@cs.ucl.ac.uk}}

\maketitle

\begin{abstract}
Vision-language models (VLMs) read an image and produce text in a single forward pass, whereas radiologists typically inspect an image several times and consult the literature before writing a report.
We introduce \framework{} (Grounded Agentic Zero-shot Evaluation), a framework that lets a medical VLM work in this iterative way by calling viewer-level tools (zoom, windowing, contrast, edge detection) and two retrieval tools backed by the U.S.\ National Library of Medicine (PubMed for medical literature, Open-i for radiological images), with structured outputs validated against a schema and full tool-call traces recorded for auditability.

On NOVA~\cite{bercea2025nova}, a benchmark of 906 brain MRI cases covering 281 rare neurological conditions, \framework{} reaches 58.2 mean average precision (mAP) at intersection-over-union (IoU) 0.3 for lesion localisation and 34.9\% Top-1 diagnostic accuracy under a joint protocol that scores captioning, diagnosis, and localisation from the image alone, without task-specific fine-tuning.
Before any tool is used, structured prompting and schema-validated outputs already improve over the published Gemini 2.0 Flash baseline (20.2 to 29.4 mAP@0.3), so framework design is itself an experimental variable.
Tool use helps rare pathologies disproportionately: the fraction of cases with IoU $> 0.3$ rises from 17\% to 58\% for diagnoses with three or fewer examples versus 25\% to 68\% for common conditions ($\geq 10$ cases), with gains tracking engagement (Gemini 3 Flash: Cohen's $d = 0.79$, 11.8 tool calls per case; Gemini 2.0 Flash: tools used in 8.2\% of cases, no significant benefit).

Retrieval ablations additionally reveal a model-dependent trade-off in which gains in diagnosis can coincide with losses in localisation, reinforcing the case for joint evaluation of diagnosis, localisation, and captioning in medical VLMs.

\end{abstract}

\section{Introduction}
\label{sec:intro}

Vision-language models (VLMs) now perform strongly on common radiology benchmarks~\cite{saab2024capabilities,yang2024advancing}, but their standard usage pattern (one image in, one text response out, in a single forward pass) differs from how radiologists actually read radiological images.
Expert readers can extract a great deal from a glance, but reliable interpretation typically involves iterative examination of multiple regions of interest and cross-referencing of findings before a diagnosis is committed to~\cite{drew2013informatics}.
This gap matters most for rare conditions, where diagnostic uncertainty is high and errors carry real clinical consequences~\cite{graber2005diagnostic}, yet current benchmarks typically compress the clinical workflow into a single forward pass.

One response to this gap is to let the VLM behave more like a reader, by calling external tools across several turns of reasoning rather than answering in one shot.
We refer to such systems as \emph{agentic}: the VLM decides, on each turn, which tool to call or whether to stop, and the final answer is the product of this sequence of decisions~\cite{bluethgen2025agentic}.
Recent agentic systems for radiology~\cite{sharma2024cxragent,medrax2025,fathi2025aura} improve over single-pass inference on chest X-ray (CXR) tasks, but are modality-specific: their tool sets consist primarily of AI-derived operations (specialist classifiers, segmentation models, counterfactual image editors) rather than the simple viewer controls a radiologist uses every day.
None integrates literature retrieval in a way that can be evaluated jointly with localisation and captioning.
Rare neurological disease on brain MRI is a particularly demanding testbed: on NOVA~\cite{bercea2025nova}, a collection of 906 cases spanning 281 rare conditions, Qwen2.5-VL-72B~\cite{bai2025qwen25vl} achieves only 24.5\% mAP@0.5 for lesion localisation and its captions cover only around 30\% of ground-truth diagnoses; RADAR~\cite{radar2025} has shown that retrieval-augmented reasoning can substantially improve diagnostic accuracy on NOVA, but reports diagnosis alone.

We introduce \framework{}, an agentic framework for medical VLMs built around two principles: the tools should match how radiologists work, and the framework should be evaluable end-to-end.
\framework{} exposes seven viewer-level visual tools that mirror everyday image-review actions (zoom, crop, contrast adjustment, window/level, edge detection, symmetry comparison, reset) together with two retrieval tools backed by the U.S.\ National Library of Medicine: \emph{search\_web} returns PubMed abstracts (guidelines, reviews, and case reports are prioritised), and \emph{search\_images} returns radiological images and captions from Open-i~\cite{demner2016openi} so that the model can compare the current study against published examples of a suspected diagnosis.
On each turn the VLM emits a structured response whose \texttt{continue} field signals whether to keep reasoning; the framework validates the response against a task-specific JSON schema, routes tool calls to the registry, and records every turn for auditability.
Task definitions supply the prompt, schema, and validators, while the adapters, tools, logging, and evaluation plumbing are reused across tasks, so \framework{} is both a tool-use controller and a stronger single-turn inference scaffold.

Our contributions are threefold.
(i)~\textbf{\framework{}}, a reusable agentic framework for medical VLMs that unifies viewer-level visual tools, PubMed and Open-i retrieval, schema-validated multi-turn outputs, and auditable tool-call traces in a single registry, with task definitions decoupled from the agent loop so the same scaffold can host other medical-VLM tasks.
(ii)~\textbf{A joint multi-task evaluation protocol for medical VLMs} that scores captioning, diagnosis, and localisation from the image alone. Under this protocol \framework{} improves substantially over the published Gemini 2.0 Flash NOVA baseline even in single-turn mode, which shows that prompting, schema-constrained outputs, and validation-aware response handling are themselves first-class experimental variables.
(iii)~\textbf{An informative retrieval failure mode} in a cross-model comparison: keyword-based PubMed retrieval can improve diagnosis while degrading localisation, with the direction and magnitude of the trade-off depending on model capability. Weaker models appear to shift bounding boxes toward the spatial patterns emphasised in retrieved abstracts, an effect that is invisible to any evaluation reporting diagnosis alone.

\section{Related Work}
\label{sec:related}

Recent agentic radiology systems share a common limitation.
CXR-Agent~\cite{sharma2024cxragent}, MedRAX~\cite{medrax2025}, and AURA~\cite{fathi2025aura} target chest X-ray and rely on AI-derived tools (specialist classifiers, segmentation models, counterfactual image editors) rather than the simple viewer controls (zoom, crop, windowing, contrast) that radiologists use daily; MedRAX, for example, reaches 63\% accuracy on ChestAgentBench by letting the model interleave reasoning and tool calls.
For volumetric data, Radiologist Copilot~\cite{yucopilot2025} adds quality-control stages to CT reporting, CT-Agent~\cite{mao2025ctagent} uses anatomy-aware tools for 3D CT question answering, and VoxelPrompt~\cite{hoopes2024voxelprompt} trains a code-generating agent for neuroimaging segmentation.
MedOpenClaw~\cite{shen2026medopenclaw} is the closest precedent: it lets VLMs drive the 3D Slicer medical-imaging viewer through a restricted interface for brain MRI and lung CT/PET, but requires a running graphical user interface (GUI).
Our tool registry exposes analogous operations as lightweight function calls, which makes batch evaluation at benchmark scale tractable.

On the retrieval side, ChestX-Reasoner~\cite{fan2025chestxreasoner} mines step-by-step reasoning chains from radiology reports and introduces a factuality-oriented reasoning metric, while RADAR~\cite{radar2025} uses dense sentence-embedding retrieval from Radiopaedia to boost GPT-4o from 24.2\% to 54.4\% Top-1 accuracy on NOVA's diagnostic task (with smaller models benefiting disproportionately).
RADAR evaluates diagnosis only.
Critically, none of these systems integrates literature retrieval into a setting where localisation and captioning are scored jointly: retrieval is typically omitted, or evaluated only against diagnosis.
\framework{} restricts retrieval to PubMed and Open-i, both maintained by the U.S.\ National Library of Medicine, and evaluates retrieval jointly with localisation and captioning.

\section{Methods}
\label{sec:methods}

\subsection{Architecture}
\label{sec:architecture}

\begin{figure}[t]
\centering
\includegraphics[width=\textwidth]{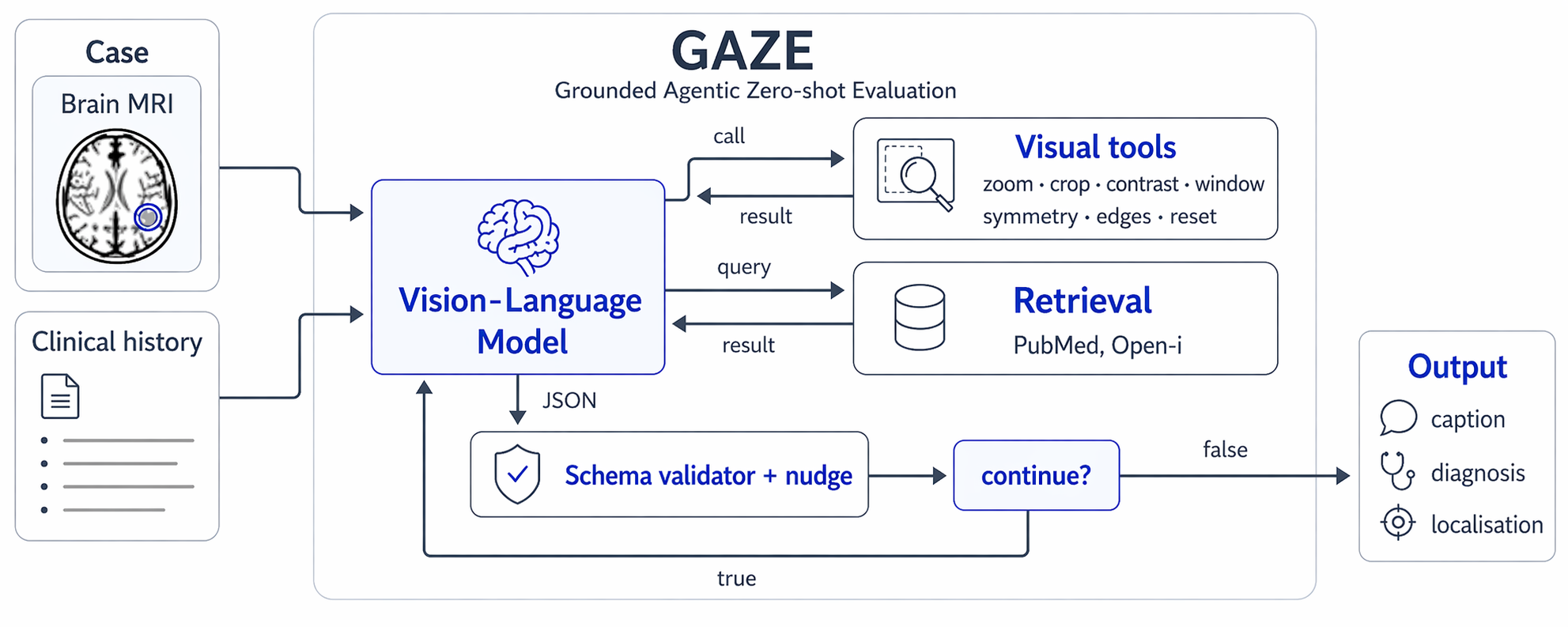}
\caption{\framework{} architecture overview. The model, not the framework, decides whether another turn is needed and which tool to call next. \framework{} validates the returned JSON against a task schema, injects nudge messages when fields are missing or malformed, and manages turn control and error recovery.}
\label{fig:architecture}
\end{figure}

At a high level, \framework{} separates task definitions from the agent machinery (Figure~\ref{fig:architecture}).
For each task, the user supplies four ingredients: a system prompt, a user message template, a JSON schema describing the expected response, and a validator that checks the model's output against the schema.
Everything else, including the conversation loop, the tool registry, response parsing, error recovery, and trajectory logging, is provided once by the framework and reused across tasks.
A thin model-adapter layer lets the same task definition run against different backends (hosted commercial APIs or local servers) without per-model changes.
The framework is task-agnostic by design: the agent loop is shared across tasks and a new benchmark requires only a task processor and evaluator.
Prompt templates, JSON schemas, and tool signatures will be released with the evaluation code.

\subsection{Tools}
\label{sec:tools}

A unified registry exposes all tools through the function-calling interface.
Visual tools return descriptive text and a base64-encoded modified image replacing the current view.

\textbf{Visual tools.}
\framework{} exposes seven tools that reproduce the everyday image-review actions used on a clinical workstation: \textit{zoom} and \textit{crop} magnify or extract a region of interest (normalised coordinates); \textit{adjust\_contrast} rescales contrast (factor 0.5--3.0); \textit{window\_level} clips the displayed intensity range at a chosen min/max to enhance specific tissue contrasts; \textit{threshold} applies a binary mask based on intensity to delineate hyperintense or hypointense lesions; \textit{symmetry\_diff} compares left and right halves to highlight asymmetric findings; \textit{detect\_edges} applies Canny edge detection; and \textit{rotate}, \textit{flip}, and \textit{reset} reorient or restore the image.
All tools include bounds checking (for example, a minimum 50-pixel crop size and zoom-factor limits) and fail closed on invalid arguments by returning a structured error while leaving the view unchanged, so the model can retry on the next turn.
Because these operations require no additional trained models, they generalise across modalities without per-modality specialist engineering.
Cropping, zooming, rotating, and flipping change the image coordinate system, so boxes produced in a modified view do not align with the coordinate frame used for scoring; the framework tracks whether any such tool has been called without a subsequent \texttt{reset}, injects a warning on the penultimate turn, and the task prompt states explicitly that un-reset coordinates yield a zero localisation score.

\textbf{Retrieval tools.}
Two retrieval tools query resources maintained by the U.S.\ National Library of Medicine.
\textit{search\_web} queries PubMed via the NCBI (National Center for Biotechnology Information) E-utilities~\cite{sayers2022eutils} and returns abstracts ranked by evidence type (guidelines, reviews, and case reports are prioritised over general articles); \textit{search\_images} queries Open-i and returns radiological images together with their captions, which lets the model compare the current study against published examples of a suspected diagnosis.
Each returned abstract is truncated to $8\,000$ Unicode characters before insertion into the prompt, which reduces the prompt-injection attack surface.

\subsection{Continuation Protocol}
\label{sec:continuation}

Each model response includes a mandatory \texttt{continue} field: if \texttt{true}, a further turn is opened with full tool access; if \texttt{false}, the response is accepted as final.
Four safeguards keep the loop bounded: (a) a configurable cap on the number of model-generation cycles (10 in all experiments), chosen up front as a practical compromise between latency, context growth, and cost; (b) a warning injected on the penultimate turn; (c) forced finalisation after three consecutive turns with no tool calls; and (d) injection of \texttt{continue: false} when the field is missing, so that a malformed response does not silently hang the loop.
The logged turn count includes intermediate tool-result steps and can therefore exceed the model-generation cap.
When the model's output fails schema validation, the framework injects a short nudge message naming the missing or malformed fields; for outputs truncated mid-JSON, it first attempts partial parsing before raising an error.

\subsection{NOVA Evaluation}
\label{sec:nova}

NOVA~\cite{bercea2025nova} contains 906 brain MRI examinations drawn from the Eurorad teaching-case archive, spanning 281 rare neurological conditions.
Annotations comprise 1{,}068 bounding boxes, produced by blinded pairwise annotation from a pool of neuroradiology residents, with senior-neuroradiologist adjudication on disagreements.
Published baseline results place Qwen2.5-VL-72B at 37.7\% mAP@0.3 and 24.5\% mAP@0.5 for lesion localisation, together with substantial loss of medical vocabulary in generated captions~\cite{bercea2025nova,bai2025qwen25vl}.

Our NOVA processor fills a prompt template with the image dimensions, clinical history, and the list of currently available tools.
The prompt instructs the model to use British English, to express bounding boxes as absolute pixel coordinates, and to report exact diagnostic names without location qualifiers.
The response schema requires a single JSON object with three top-level sections (caption, diagnosis, localisation); the validator clamps out-of-range confidences and verifies that bounding-box coordinates are well formed.

\textbf{Metrics.}
Localisation is reported as per-image mean average precision (mAP) at three intersection-over-union (IoU) thresholds.
IoU measures the overlap between a predicted and a ground-truth bounding box: the area of intersection divided by the area of union.
mAP@0.3 and mAP@0.5 require at least 30\% and 50\% overlap respectively; mAP@[0.5:0.95] averages mAP over ten IoU thresholds from 0.50 to 0.95 in steps of 0.05.
We compute mAP with 11-point interpolation~\cite{everingham2010pascal}; because the NOVA evaluation code is not public, we use this implementation throughout and note comparability caveats explicitly.
Caption quality is measured with METEOR~\cite{banerjee2005meteor} and Clinical F1 (a keyword-based F1 over clinical terms, following~\cite{bercea2025nova}).
Diagnosis is evaluated with Top-1 accuracy (the fraction of cases where the model's top prediction matches the ground-truth label) and Top-5 accuracy (the fraction where the ground-truth label appears in the model's top five candidates), using GPT-4o as an LLM-as-judge for semantic matching against ground-truth labels.

Critically, we evaluate all three tasks jointly from the image alone, whereas the published NOVA baselines evaluate each task independently and supply the ground-truth caption as an oracle input for diagnosis.
The joint protocol is harder and more clinically realistic.

\section{Experiments}
\label{sec:experiments}

\subsection{Setup}

We evaluate five configurations on all 906 NOVA cases, added incrementally to isolate the contribution of each framework component:
(1)~\textbf{Single-turn}: one forward pass, no tools;
(2)~\textbf{+Reasoning}: multi-turn loop, but still no tools;
(3)~\textbf{+Search}: multi-turn loop with PubMed and Open-i retrieval only;
(4)~\textbf{+Visual}: multi-turn loop with viewer-level visual tools only;
(5)~\textbf{Full}: multi-turn loop with all tools and retrieval.

For the same-model comparison we use Gemini 2.0 Flash, the same model family reported in the NOVA baseline~\cite{bercea2025nova}, so that any observed difference can be attributed to the framework rather than to a change of model.
This comparison captures two distinct effects: the difference between the published baseline and our single-turn result reflects \framework{} as a stronger single-turn inference scaffold, while the differences among our own configurations isolate the incremental effect of multi-turn reasoning, retrieval, and viewer-level tools.

We then evaluate across five models spanning two families: Gemini 2.0 Flash, Gemini 3 Flash, and three variants of Qwen3.5~\cite{qwen2026qwen35} at sizes denoted 35B-A3B, 122B-A10B, and 397B-A17B.
Qwen3.5 is a mixture-of-experts (MoE) model family: the first number gives the total parameter count, while the \textsc{A} suffix gives the number of parameters active for a given forward pass (for example, 397B-A17B has 397\,B parameters in total but only 17\,B active per token).
Model identifiers, routing metadata, and prompts are recorded in the released configs; Gemini 3 Flash had no public model card at run time.
All models are accessed via OpenRouter at temperature 0.0, seed 42, maximum 10 turns, and native $480 \times 480$ input resolution (images are not resized); experiments were run in March--April 2026.

We evaluate diagnosis by LLM-as-judge semantic matching with GPT-4o (OpenRouter model ID \texttt{openai/gpt-4o}), following the NOVA protocol~\cite{bercea2025nova}; the study uses only the publicly available, de-identified NOVA dataset.
Our primary comparison is against the zero-shot baselines in~\cite{bercea2025nova} because they report all three tasks (captioning, diagnosis, localisation), whereas RADAR~\cite{radar2025} reports diagnosis only.
For the oracle-caption diagnostic task we follow NOVA Task 3 (text-only diagnosis from the ground-truth caption), which also matches RADAR's evaluation setting.

Two caveats apply throughout.
The NOVA paper describes ``COCO-style'' mAP without specifying global pooling or per-image averaging; our per-image implementation reproduces the published baseline magnitudes most closely, but exact metric equivalence cannot be guaranteed.
Provider-specific token costs vary over time and across routing backends, so we report turns and tool calls per case as provider-agnostic proxies for test-time compute, and single-turn versus multi-turn comparisons should be read as combining changes in inference scaffold, tool use, and test-time compute allocation.

\subsection{Results}

\begin{table}[t]
\centering
\caption{Same-model comparison on NOVA using Gemini 2.0 Flash. The first block reports the published NOVA baseline; the second block reports our single-turn reimplementation and \framework{} ablations under a joint all-in-one evaluation protocol. MET = METEOR; C-F1 = Clinical F1; T-1/T-5 = Top-1/Top-5 diagnostic accuracy; Fail = run-level failures excluded from metric means. Bold = best within each metric column (excluding the published baseline). \ddag~Oracle-caption protocol (text-only diagnosis, matching NOVA Task 3).}
\label{tab:ablation}
\setlength{\tabcolsep}{3.5pt}\small
\begin{tabular}{@{}l ccc cc cc r@{}}
\toprule
& \multicolumn{3}{c}{\textbf{Localisation}} & \multicolumn{2}{c}{\textbf{Caption}} & \multicolumn{2}{c}{\textbf{Diagnosis}} & \\
\cmidrule(lr){2-4} \cmidrule(lr){5-6} \cmidrule(lr){7-8}
Config & @0.3 & @0.5 & @[0.5:0.95] & MET & C-F1 & T-1 & T-5 & Fail \\
\midrule
\multicolumn{9}{@{}l}{\textit{Published NOVA baseline}} \\
NOVA (reported) & 20.2 & 7.4 & 2.0 & 15.2 & 19.8 & 22.1 & 37.4 & -- \\
\midrule
\multicolumn{9}{@{}l}{\textit{\framework{} (ours)}} \\
Single-turn & 29.4 & 10.9 & 3.0 & 17.8 & 20.1 & 16.9 & 23.7 & 21 \\
+Reasoning  & 27.6 & 8.9  & 2.6 & 18.0 & 20.0 & 13.5 & 21.3 & \textbf{0} \\
+Search     & \textbf{32.4} & 11.0 & \textbf{3.4} & 18.3 & 19.9 & 14.3 & 22.2 & 32 \\
+Visual     & 31.2 & \textbf{12.0} & 3.3 & 18.3 & 20.0 & 14.2 & 21.9 & \textbf{0} \\
Full        & 31.2 & 10.9 & 2.8 & \textbf{18.4} & \textbf{20.2} & 13.3 & 20.8 & \textbf{0} \\
\midrule
Oracle\ddag & -- & -- & -- & -- & -- & \textbf{33.4} & \textbf{44.0} & 0 \\
\bottomrule
\end{tabular}
\end{table}

\begin{table}[t]
\centering
\caption{Cross-model view of the retrieval trade-off on NOVA (906 attempted cases). We report the primary spatial metric and diagnosis scores; Fail denotes run-level failures excluded from metric means. The key pattern is not a single best configuration, but that search can move diagnosis and localisation in opposite directions. Bold = best within each model block.}
\label{tab:models}
\setlength{\tabcolsep}{4pt}\small
\begin{tabular}{@{}ll ccc r@{}}
\toprule
& & \textbf{Localisation} & \multicolumn{2}{c}{\textbf{Diagnosis}} & \\
\cmidrule(lr){3-3} \cmidrule(lr){4-5}
Model & Config & mAP@0.3 & Top-1 & Top-5 & Fail \\
\midrule
\multicolumn{6}{@{}l}{\textit{Published NOVA baselines}} \\
Gemini 2.0 Flash & Single & 20.2 & 22.1 & 37.4 & -- \\
Qwen2.5-VL-72B   & Single & 37.7 & 22.4 & 35.2 & -- \\
\midrule
\multicolumn{6}{@{}l}{\textit{Gemini 3 Flash (\framework{}, ours)}} \\
& Single  & 23.9 & 32.2 & 45.3 & 308 \\
& +Reason & 29.5 & 33.0 & 41.9 & 2 \\
& +Search & 26.6 & 32.3 & 43.0 & 19 \\
& +Visual & 46.8 & 31.9 & 42.9 & 8 \\
& Full    & \textbf{58.2} & \textbf{34.9} & \textbf{46.6} & 13 \\
\midrule
\multicolumn{6}{@{}l}{\textit{Qwen3.5-122B-A10B (\framework{}, ours)}} \\
& Single  & 39.1 & 20.4 & 28.7 & 0 \\
& +Reason & 38.5 & 17.9 & 28.0 & 0 \\
& +Search & 20.9 & \textbf{23.3} & \textbf{36.0} & 0 \\
& +Visual & 42.1 & 19.3 & 32.1 & 0 \\
& Full    & \textbf{43.2} & 19.8 & 32.5 & 0 \\
\midrule
\multicolumn{6}{@{}l}{\textit{Other Qwen3.5 full configurations (\framework{}, ours)}} \\
Qwen3.5-35B-A3B   & Full & 48.2 & 17.9 & 29.7 & 0 \\
Qwen3.5-397B-A17B & Full & 57.6 & 22.3 & 33.5 & 0 \\
\bottomrule
\end{tabular}
\end{table}

Table~\ref{tab:ablation} compares configurations for a single model (Gemini 2.0 Flash).
In the single-turn regime, \framework{} improves over the published baseline on both localisation and captioning (mAP@0.3: 20.2 to 29.4; METEOR: 15.2 to 17.8), a direct consequence of structured prompting, schema-constrained outputs, and validation-aware response handling rather than of any tool use.
Within our own configurations, +Search gives the best mAP@0.3 (32.4), +Visual gives the best mAP@0.5 (12.0), and Full gives the best caption score (18.4 METEOR), with +Search also producing the most run failures (32 vs.\ 21).
The joint all-in-one diagnostic accuracy (16.9\% Top-1) is lower than the published NOVA baseline (22.1\%) because NOVA supplies the ground-truth caption as oracle input to the diagnostic task; when we supply the same oracle caption, \framework{} reaches 33.4\% Top-1 (a 51\% relative improvement).
This has a methodological implication: the framework itself can change measured performance before any external tool is used.

Table~\ref{tab:models} compares across model families.
Gemini 3 Flash with full tool access attains the best scores, at 58.2 mAP@0.3 and 34.9\% Top-1 accuracy under the joint protocol.
The open-weight Qwen3.5-397B-A17B reaches 57.6 mAP@0.3 with tools, nearly matching Gemini 3 Flash.
At the smallest active-parameter scale, Qwen3.5-35B-A3B surpasses the published Qwen2.5-VL-72B localisation baseline on NOVA (48.2 vs.\ 37.7 mAP@0.3), although the comparison should be read alongside the very different total-to-active parameter ratio of the MoE models.
Retrieval should be evaluated as a multi-task intervention, because the same retrieval call that improves diagnosis can simultaneously degrade localisation (see Analysis).

\begin{figure}[t]
\centering
\includegraphics[width=0.85\textwidth]{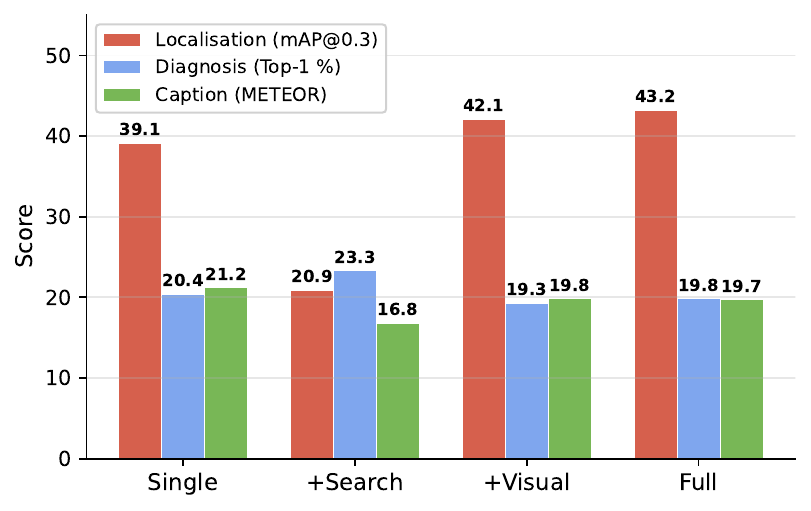}
\caption{\textbf{Retrieval trade-off for Qwen3.5-122B-A10B.} +Search improves diagnosis but reduces localisation by $\approx$47\%; visual tools recover localisation. A diagnosis-only evaluation would miss this spatial regression.}
\label{fig:dissociation}
\end{figure}

\subsection{Analysis}

\textbf{Run failures.}
We report failed runs (cases for which no valid output was produced) separately and exclude them from metric means.
For Gemini 3 Flash, the failures shown are provider- and model-routing errors during the evaluation window rather than schema-validation failures on the model side.
The reduction from 308 single-turn failures to 13 full-configuration failures therefore reflects provider availability, not any capability of the framework or the model.

\textbf{Tool benefit tracks model engagement.}
Paired permutation tests on per-sample IoU show that the localisation benefit of tools varies substantially across models: Gemini 3 Flash shows a large effect (Cohen's $d = 0.79$, $p < 10^{-4}$), Qwen3.5-397B-A17B and Qwen3.5-35B-A3B show small effects ($d = 0.25$ and $d = 0.12$, both $p < 10^{-3}$), and Gemini 2.0 Flash shows no benefit ($d = -0.03$, $p = 0.32$).
This gradient reflects both willingness and ability to engage the tool interface.
Gemini 2.0 Flash uses tools in only 8.2\% of cases (0.4 calls per successful case), whereas Gemini 3 Flash and the Qwen3.5 family engage tools in nearly all successful full-configuration cases, at 6.1--11.8 calls over 12.3--19.0 logged turns (Qwen3.5-122B-A10B sits at the lower end, with 6.1 calls over 12.3 turns).
Across all models the three most frequently called tools are \texttt{reset}, \texttt{window\_level}, and \texttt{zoom}.
For Gemini 2.0 Flash, multi-turn reasoning without tools gives no significant localisation benefit ($p = 0.32$).
For Gemini 3 Flash, the raw reasoning-only gain (23.9 to 29.5) is not significant either when restricted to samples that succeeded in both conditions ($p = 0.43$), which is consistent with the large tool-augmented gains coming from tool interaction rather than from additional generation turns.

\textbf{Search trade-off is model-dependent.}
Adding retrieval produces very different effects across models.
For Qwen3.5-122B-A10B, search reduces localisation by $46.5\%$ (mAP@0.3: 39.1 to 20.9) while improving diagnosis (Top-1: 20.4\% to 23.3\%, Top-5: 28.7\% to 36.0\%; see Figure~\ref{fig:dissociation}).
Gemini 3 Flash shows only a modest localisation gain (23.9 to 26.6) with negligible diagnosis change (32.2\% to 32.3\%), which suggests that stronger models can integrate textual retrieval without spatial harm.
Gemini 2.0 Flash shows the opposite pattern (localisation 29.4 to 32.4, diagnosis 16.9\% to 14.3\%).
We do not run a formal Model $\times$ Search interaction test; the three-model pattern is descriptive evidence that the trade-off varies across models.
The trade-off is therefore not an intrinsic property of keyword retrieval but depends on whether the model can correctly weight textual evidence against what it sees in the image.
When a weaker model receives abstracts describing several conditions with overlapping terminology but different spatial presentations (for example, multiple sclerosis plaques, tumefactive MS, and lymphoma all affecting the corpus callosum), it appears to shift its bounding box toward the most common spatial pattern in the retrieved text rather than the lesion visible in the image.

One representative case illustrates the mechanism.
In a NOVA case of MELAS (mitochondrial encephalomyopathy with stroke-like episodes), Qwen3.5-122B-A10B in single-turn mode predicts glioblastoma yet achieves IoU $= 0.68$ because its bounding box still captures the hyperintense cortical lesion in the image.
When search is added, the model retrieves PubMed abstracts describing characteristic cortical and subcortical distributions of MELAS, correctly revises its diagnosis to MELAS, but shifts its bounding box toward the subcortical white-matter pattern emphasised in the retrieved text, collapsing IoU to $0.05$.
The diagnosis improved; the localisation degraded because textual evidence overrode what the model saw in the image.

Visual tools show a different pattern in isolation: they improve localisation for Gemini 3 Flash ($+22.9$ mAP@0.3), Qwen3.5-122B-A10B ($+3.0$), and Gemini 2.0 Flash ($+1.8$), while slightly degrading diagnosis.
For Qwen3.5-122B-A10B, the full configuration recovers localisation but gives back part of the diagnostic gain seen with search alone.
For Gemini 3 Flash, the combined improvement ($+34.3$) exceeds the sum of search ($+2.7$) and visual ($+22.9$), consistent with a super-additive interaction between the two kinds of tools.
Retrieval systems for medical VLMs should therefore be evaluated on all tasks they might affect, not diagnosis alone.

\textbf{Rarity stratification.}
For Gemini 3 Flash, tool augmentation improves localisation across rarity tiers when comparing Single to Full, with a larger relative gain on rare conditions.
For pathologies with at most three cases in the benchmark (36\% of samples), the fraction of cases with IoU $> 0.3$ rises from 17\% to 58\% (a 241\% relative gain, with run failures counted as misses).
For common conditions ($\geq 10$ cases), the same comparison gives 25\% to 68\% (172\% relative).

\section{Discussion}
\label{sec:discussion}

\textbf{The framework itself is an experimental variable.}
The gap between the published Gemini 2.0 Flash NOVA baseline and our single-turn configuration should not be attributed to tool use: task-specific prompting, schema-constrained outputs, and validation-aware response handling are themselves variables worth reporting.
A medical-agent framework is therefore best understood as a systems layer for task specification, validation, trajectory logging, and evaluation, not merely as a wrapper around tools.
The same interface can host other medical-VLM tasks (visual question answering, report generation, visual grounding) by swapping task processors while reusing the adapters, tools, traces, and evaluation plumbing.

\textbf{Joint versus oracle protocols.}
Our joint-protocol diagnostic accuracy (16.9\% Top-1) is lower than the published NOVA baseline (22.1\%) because NOVA supplies the ground-truth caption as an oracle input to diagnosis; when we supply the same oracle caption, \framework{} reaches 33.4\% Top-1, a 51\% relative improvement.
The joint protocol is harder and more clinically realistic, because oracle inputs can mask failure modes that matter in practice.

\textbf{Tool benefit follows engagement.}
Tool benefit is better understood as a spectrum of integration quality than as an all-or-nothing capability.
Gemini 2.0 Flash uses tools in only 8.2\% of cases and derives no measurable benefit, whereas Gemini 3 Flash achieves the largest gain ($d = 0.79$) with 11.8 calls per case.
The Qwen3.5 35B and 397B variants engage heavily (9.0--11.0 calls) yet show only small gains ($d = 0.12$--$0.25$), so call frequency alone is not sufficient.
Benchmarks should therefore report tool engagement alongside task accuracy.

\textbf{Retrieval design.}
RADAR~\cite{radar2025} reported higher diagnostic accuracy on NOVA using Radiopaedia retrieval.
We restricted \framework{}'s retrieval to PubMed and Open-i, both of which are explicitly distributed by the U.S.\ National Library of Medicine for biomedical research, and did not use Radiopaedia: at the time of our evaluation, Radiopaedia's published access policy did not permit unrestricted automated AI use without prior authorisation,\footnote{Radiopaedia's \texttt{robots.txt} declares \texttt{ai-train=no} and the publicly listed Terms of Use ask that the site not be used as an imaging dataset for machine learning or artificial-intelligence research without permission.} so we preferred sources whose terms aligned with the open-research framing of this work.
The gap in absolute scores is consistent with two methodological differences: Radiopaedia is curated specifically as a radiology teaching resource and likely has a higher signal-to-noise ratio for NOVA's rare-disease diagnoses than general PubMed abstracts, and RADAR uses dense embedding-based retrieval rather than keyword search.
For reference, RADAR reports Gemini 2.0 Flash at 45.6\%/66.2\% Top-1/Top-5 without retrieval and 48.5\%/72.1\% with Radiopaedia retrieval under the same oracle-caption protocol~\cite{radar2025}, compared to our oracle Top-1/Top-5 of 33.4\%/44.0\% (Table~\ref{tab:ablation}).
Denser, more spatially grounded retrieval~\cite{karpukhin2020dense,xiong2021approximate} from open-licensed sources is a natural next step, but will still require models that can resolve conflicts between textual and visual evidence.

\textbf{Auditability and compliance.}
\framework{}'s trace-logging is designed with regulatory requirements for agentic medical AI in mind~\cite{bluethgen2025agentic}: logged tool traces support record-keeping obligations of the kind described in EU AI Act Article~12~\cite{euaiact2024}, coordinate-space modifications are tracked through tool invocations, and the prompt-injection surface is reduced by truncating retrieved abstracts before they enter the prompt.
Related work under the MHRA AI Airlock pilot reports that retrieval-augmented generation can reduce hallucinations in a guideline-grounded medical question-answering sandbox~\cite{mhraairlock2025}, although that use case differs from ours.

\textbf{Limitations.}
We evaluate on brain MRI only, with 2D visual tools, and single-turn versus multi-turn comparisons are confounded by test-time compute (higher latency and token cost, with provider routing that prevents precise cost attribution).
The comparison with the published NOVA baseline aggregates differences in prompt design, schema constraints, OpenRouter routing, and possibly the mAP implementation itself, so the improvement should not be attributed solely to framework architecture.
We also cannot guarantee that NOVA was excluded from the pre-training corpora of hosted models, nor have we measured inter-rater agreement between the GPT-4o judge and human raters.

\textbf{Future work.}
We plan to extend \framework{} to medical visual question answering, report generation, and visual grounding, and to use the logged trajectories for reinforcement-learning-style optimisation of tool policies and reasoning-quality metrics such as RadRScore~\cite{fan2025chestxreasoner}.

\section{Conclusion}
\label{sec:conclusion}

\framework{} combines structured prompting, schema-constrained outputs, multi-turn reasoning, viewer-level visual tools, and literature retrieval into a single framework for medical VLM evaluation.
On NOVA, it reaches 58.2 mAP@0.3 and 34.9\% Top-1 accuracy under joint all-in-one evaluation without task-specific fine-tuning, improves over the published Gemini 2.0 Flash localisation baseline even in single-turn mode, and delivers larger relative gains on rare pathologies than on common ones.
Within this picture we identify one informative failure mode, a model-dependent retrieval trade-off in which keyword-based search can move diagnosis and localisation in opposite directions, which argues for joint multi-task evaluation whenever retrieval is introduced into a medical VLM pipeline.
Code, prompt templates, schemas, and tool definitions will be released upon publication.

\begin{credits}
\subsubsection{\discintname}
The authors have no competing interests to declare.
\end{credits}

\clearpage
\bibliographystyle{splncs04}
\bibliography{references}

\end{document}